\documentclass[10pt, sigconf]{acmart}

\usepackage{graphicx}
\usepackage{multirow}
\usepackage{algorithm}
\usepackage[noend]{algpseudocode}
\usepackage[english]{babel}
\usepackage{appendix}
\usepackage{enumitem}
\usepackage{amssymb}
\usepackage{amsmath}
\usepackage{color}

\setcopyright{none}
\renewcommand\footnotetextcopyrightpermission[1]{}

\newcommand{\name} {{HawkEye}}

\newcommand{\colorize} {\color{black}}

\begin{document}

\title{
High Resolution Millimeter Wave Imaging For Self-Driving Cars
}

\newcommand{\supsym}[1]{\raisebox{6pt}{{\footnotesize #1}}}


\author{Junfeng Guan, Sohrab Madani, Suraj Jog, Haitham Hassanieh}
\affiliation{%
	\institution{ Univeristy of Illinois at Urbana Champaign}
\and \{jguan8, smadani2, sjog2, haitham\}@illinois.edu}

\date{}
\maketitle

\newcommand*\wrapletters[1]{\wr@pletters#1\@nil}
\def\wr@pletters#1#2\@nil{#1\allowbreak\if&#2&\else\wr@pletters#2\@nil\fi}
\newcommand\ie{\textit{i.e.}}
\newcommand\newchanges[1]{{#1}}

\pagestyle{plain}
\thispagestyle{plain}

\setlength{\abovedisplayskip}{3pt}
\setlength{\belowdisplayskip}{3pt}

\begin{sloppypar}

	\noindent{\bf Abstract --}
Recent years have witnessed much interest in expanding the use of networking signals beyond 
communication to sensing, localization, robotics, and autonomous systems.  This
paper explores how we can leverage recent advances in 5G millimeter wave (mmWave) technology 
for imaging in self-driving cars. Specifically, the use of mmWave in 5G has
led to the creation of compact phased arrays with {\colorize hundreds} of antenna elements that
can be electronically steered. Such phased arrays can expand the use of mmWave
beyond vehicular communications and simple ranging sensors to a full-fledged imaging {\colorize system}
that enables self-driving cars to see through fog, smog, snow, etc. 

Unfortunately, using mmWave signals for imaging in self-driving cars is challenging
due to the very low resolution, the presence of fake artifacts 
resulting from multipath reflections and the absence of portions of the car due
to specularity. This paper presents \name, a system that can enable high
resolution mmWave imaging in self driving cars. \name\ addresses the above
challenges by leveraging recent advances in deep learning known as 
Generative Adversarial Networks (GANs). \name\ introduces a
 GAN architecture that is customized to mmWave imaging and builds a system 
that can significantly enhance the quality of mmWave images for self-driving 
cars.


	\section{Introduction}

Self-Driving cars have generated a great deal of excitement.  Last year, 
Tesla released a more affordable car model that supports automatic driving on
highways and city streets and can respond to traffic lights and stop signs with
a software update. More than 140,000 of these cars are already on the
road~\cite{Tesla1}. Another company, Waymo, recently launched the first
commercial self-driving taxi service in Arizona~\cite{waymo} and major car
manufacturers like Toyota and Honda have invested up to $\$3$ billion in
research and development on autonomous driving~\cite{toyota, honda,
investment}.  

Despite this progress, the vision of fully autonomous vehicles, referred to as
level 5 in the standards for driving automation, still remains a distant
goal~\cite{mit_uber, mit_self_driving, regulator}. One of the major obstacles
that the technology must overcome is its inability to function in inclement
weather conditions such as fog, smog, snowstorms and sandstorms~\cite{sleet,
headache, foggy}.  Autonomous vehicles primarily use LiDARs or cameras to
obtain an accurate and reliable view of the environment. However, since these
sensors rely on optical frequencies for imaging, they suffer in low visibility
conditions such as those caused by fog, snow and dust
particles~\cite{Laurenzis2011,Satat2016,Satat}. Cameras also suffer at night in low light
conditions. This is problematic as many manufacturers including Tesla avoid
using LiDAR altogether, {\colorize making cameras their primary sensory module}~\cite{Tesla}.

Millimeter wave (mmWave) wireless signals, on the other hand, offer far
more favorable characteristics due to their ability to work at night and in
inclement weather~\cite{RobustTI,EMatten}. This is why car manufacturers today use
mmWave radar for the purpose of ranging, i.e., to determine the distance to
other vehicles~\cite{audiRadar}. These radars, however, are limited
to unidirectional ranging and cannot fully image the environment~\cite{BOSCHACC}. 
Luckily, the advent of 5G networks in the mmWave spectrum has led
to the creation of compact phased arrays with hundreds of antennas~\cite{Bell,UCSD}. Such technology enables 
extremely narrow beams which can be steered electronically.

In this paper, we explore whether such 5G mmWave technology can extend
beyond communication to enable full-fledged imaging in self-driving cars.
In particular, by electronically steering the beam, one can capture
reflections from different regions in space and potentially image
the environment. Unfortunately, simply using very narrow
steerable beams is not sufficient for three reasons:

\begin{figure*}[t!]
\centering
\includegraphics[width=6.5in]{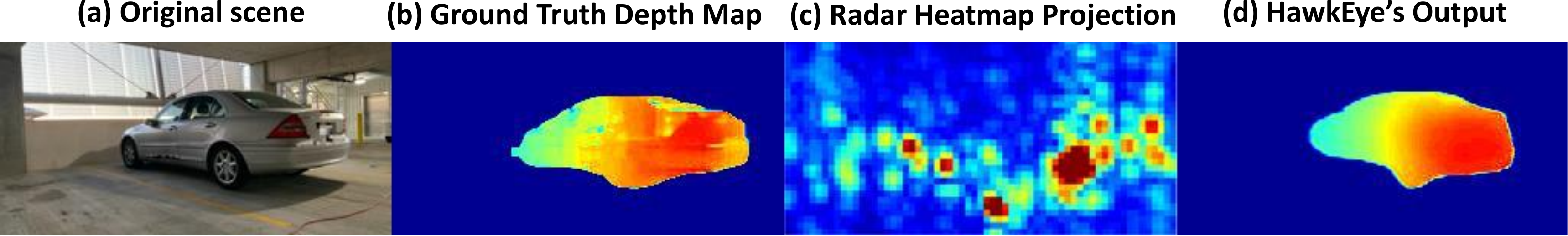}
\vskip -0.1in
\caption{\footnotesize Specularity and Low Resolution in mmWave radar results in poor perceptual quality for images. To overcome this limitation, we exploit GAN architectures to create high resolution images from the low resolution mmWave radar heatmaps. The output from \name\ is comparable to that captured from a stereo camera system.}
\label{fig:teaser}
\vskip -0.2in
\end{figure*}

\begin{itemize}[leftmargin=*,topsep=6pt]
	\setlength{\itemsep}{0pt} 
\setlength{\parskip}{0pt}
\item {\bf Resolution:} While mmWave does offer higher resolution
	than standard wireless technologies, the spatial resolution is nowhere near
		what can be obtained from LiDARs or cameras. 
		Fig.~\ref{fig:teaser}(c) shows an example of an image
		obtained from a high resolution mmWave system. The image
		appears as blobs of RF reflections and carries little
		to no contextual and perceptual information as opposed
		to a camera image and its corresponding 2D depth map shown 
		in Fig.~\ref{fig:teaser}(a,b). 

\item {\bf Specularity:} Wireless mmWave signals are highly specular i.e., the
	signals exhibit mirror-like reflections from the car~\cite{Specular1}. As a
		result, not all reflections from the car propagate back to the 
		mmWave receiver and major parts of the car do not appear in the image.
		This makes it  almost impossible to detect the shape, size, and 
		even orientation of the car as can be seen in Fig.~\ref{fig:teaser}(c). 
		
\item {\bf Multipath:} Wireless reflections from the car can also bounce off the road and other cars
	and travel along multiple paths to the mmWave receiver
	creating shadow reflections in various locations in the scene as
		shown in Fig.~\ref{fig:teaser}(c). Hence, multipath can result
		in noisy artifacts in the image that are hard to interpret or
		dismiss as fictitious obstacles.

\end{itemize}

To address the above challenges, today's commercial mmWave imaging systems,
like airport scanners, use human-sized mechanically steerable arrays to improve the resolution.  They
also isolate the object being imaged in the near field to eliminate multipath and rotate the
antennas {\colorize around the object} to address specularity~\cite{Mamandipoor2015,Board2017}.  However, such
a design would be extremely bulky and not practical for self-driving cars as we
have no control over the cars being imaged.

In this paper, we introduce \name, a system that can significantly improve the
quality of mmWave imaging for self-driving cars by generating high resolution,
accurate and perceptually interpretable images of cars from 3D mmWave heat-maps
as shown in Fig.~\ref{fig:teaser}(d). To achieve this, \name\ builds on a recent
breakthrough in machine learning referred to as  GANs (Generative
Adversarial Networks). GAN is a learning paradigm that has
proved to be a very effective tool for improving image resolution and
generating realistic looking images~\cite{goodfellow2014generative,Ledig2017}. \name\ trains a GAN using
stereo depth maps obtained from camera images (as ground truth) to learn a mapping that
increases the resolution, fills in the missing gaps due to specularity and
eliminates artifacts caused by multipath.  This allows \name\ to estimate the
location, shape, orientation, and size of the car in the scene.

Designing and training a GAN architecture that is capable of transforming 3D
mmWave heatmaps to high resolution 2D depth maps poses new
challenges beyond standard vision-based learning for image-to-image
translation\footnote{This includes increasing resolution of a camera image,
transforming black and white images to colored, transforming cartoon images
to photo-realistic images, generating images with modified context,
etc.~\cite{pix2pix,Cycle-GAN,Hassan2018}}. Specifically, \name\ addresses
the following three challenges in adapting GANs to mmWave imaging.

\vskip 6pt \noindent {\bf 1) Discrepancy in Perspective \& Dimensions:} \name\
must learn to generate a high resolution 2D depth-map from a 3D mmWave
heat-map. That is, \name's ground-truth and output correspond to a 2D image
with pixels corresponding to depth, while its input corresponds to a 3D image
where every voxel value maps to RF signal strength. However, input and output in GANs 
are usually both 2D \cite{yoo2016pixel,zhu2017toward,zhang2017image} or 
both 3D ~\cite{jin2018ct}. One solution to this discrepancy would be to
transform the 2D depth maps to 3D point clouds where each voxel value maps to a
binary one or zero, and use a point cloud GAN similar to~\cite{PointGAN} to
generate 3D point clouds from 3D heat-maps. However, such a design results in
very sparse high-dimensional {\colorize output data}. Training such a 3D GAN is
known to be notoriously hard. It requires the network to learn significantly
more parameters which exponentially increases the {\colorize optimization} search space making it
difficult for the network to converge~\cite{smith2017improved}. Instead, \name\ separately encodes 
both the 2D input and 3D output into a common 1D space through
separate neural networks (Fig.~\ref{fig:g}). This significantly reduces the complexity of 
the network~\cite{badrinarayanan2017segnet} while also allowing
for the direct comparison of encoded input and output, as will be
discussed in detail in section \ref{sec:gan}.

\vskip 6pt \noindent {\bf 2) Quantitative Accuracy:}
GANs have proven to be an extremely powerful tool for generating
realistic images that are hard to distinguish from real life
images~\cite{DBLP:journals/corr/abs-1812-04948}. However, \name's goal is not simply to generate a realistic
looking car, but to also reconstruct the accurate shape, size, orientation, and
location of the car from the mmWave heat-map.  In order to ensure quantitative
accuracy, we modify the loss function that the GAN aims to
optimize. Specifically, instead of simply using the standard GAN loss, we add
an L1\footnote{L1 loss represents the absolute value of the difference between
pixels in the output and ground-truth.} loss term and a perceptual loss term
to ensure the output is quantitatively and perceptually close to the
ground-truth. 


In addition to modifying the loss function, we also adapt the idea of 
{\it skip-connections} to our GAN. Specifically, {\it skip-connections} allow us
to directly map features between the input and the output which are typically
difficult for GANs to learn~\cite{ronneberger2015u}. For example, mmWave provides very
accurate ranging data which should be mapped directly to the output. Without
such {\it skip-connections}, the GAN might degrade the accuracy of the car's
location. \name\ adapts {\it skip-connections} to directly transfer 
such features from 3D input to 2D output.

\vskip 6pt \noindent{\bf 3) Training:} Training GANs requires collecting
thousands of data samples to ensure that the system generalizes well and does
not overfit to the training data.  However, collecting such a huge number of
mmWave images on real cars takes a prohibitively long time. To address this, we
start by training our GAN using synthesized mmWave images. We build a simulator
to generate {\colorize mmWave heatmaps} from 3D CAD models of various car types. The simulator
uses precise ray tracing models to capture the propagation and reflection
characteristics of mmWave RF signals as well as specularity and artifacts
resulting from multipath. After training with synthesized data, we fine tune
the GAN using real data that we captured in order to create a more robust and
reliable system. Finally, we test on real data that was not included in the
training set using standard $k$-fold cross-validation.



\vskip 6pt \noindent{\bf Implementation \& Evaluation:}
We implement \name\ and empirically
evaluate its performance on real mmWave images. We build a mmWave imaging {\colorize  module } that uses FMCW waveform at 60 GHz and Synthetic Aperture Radar (SAR) to emulate a large
antenna array {\colorize  for capturing 3D mmWave heatmaps of the scene}. We collect 240 real images of 40 different car models from various viewpoints, distances and backgrounds. {\colorize To evaluate \name's performance under different weather conditions, we also collect images in various visibility conditions ranging from clear visibility to different degrees of fog and rain.}
We summarize the results as follows:

\begin{itemize}[leftmargin=*,topsep=3pt]
	\setlength{\itemsep}{0pt} 
\setlength{\parskip}{0pt}

\item {\bf Qualitative Results:} We provide several
	qualitative results similar to Fig.~\ref{fig:teaser}
		in section~\ref{sec:qual}. The results show
		that \name\ is able to accurately recover the {\colorize shape, size and orientation}
		of the car from the mmWave input. {\colorize We also show \name's performance in low visibility conditions like fog and rain, and
		demonstrate that \name\ is able to recreate accurate images in such low visibility conditions and can generalize to different environments.}


\item {\bf Quantitative Results:} We evaluate \name\ by computing
	the error in its estimates of the location, orientation, 
	shape, and size of the car. 
		Our results show that \name\ improves the accuracy
		of these estimates $2.2 - 5.5\times$ compared 
		to the 3D mmWave heatmaps . 

\end{itemize}

\vskip 4 pt
\noindent{\bf Contribution: } This paper has the following contributions:

\begin{itemize}[leftmargin=*,topsep=3pt]
	\setlength{\itemsep}{0pt} 
\setlength{\parskip}{0pt}
\item The paper presents the first system that can 
	construct high resolution 2D depth maps of
	cars from noisy low resolution mmWave images.
	The system consists of four main
	components: a simulator that generates
	synthetic training data, a mmWave imaging system that
	captures real data, deep learning GAN architecture that 
	reconstructs the output images, and a system for extracting
	quantitative measures such as size and orientation
	from the output. 

\item The paper presents a new GAN architecture that
	is tailored for mmWave imaging to account for
		perspective mismatch and quantitative
		accuracy. 

\item The paper demonstrates a working system that provides significant improvement
	in the quality of mmWave imaging for self-driving cars. {\colorize It also demonstrates that the system can work 
	in low visibility conditions like fog and rain through real test data points collected in such inclement weather conditions.}

\end{itemize}

\vskip 4 pt

\noindent{\bf Limitations: } 
HawkEye presents the first step towards high
resolution imaging with mmWave wireless systems. However, 
HawkEye has a few limitations that are worth discussing. 
Notably, due to the lack of cheap, commercial phased arrays, 
our experimental setup uses SAR which precludes 
real-time frame capture. As phased
array technology becomes more accessible in the coming
years [49, 72], we plan to extend our platform to a real-time system. 
Another limitation is that our GAN was mainly trained on cars.
We plan to extend it to pedestrians, bicycles and other road infrastructure in the future. We elaborate on these limitations in
Section 10 and we propose future directions to address them.

	\section{Related Work}

\noindent{\bf (a) Sub-6 GHz Wireless Imaging Systems:} Our paper is related
to recent work on sub-6 GHz wireless imaging of the human body.
The work aims to extract and track the human skeleton to estimate
the 3D pose of the body~\cite{RF-Capture, RF-Pose3D,RF-Pose}. 
It leverages human motion to combat specularity by combining reflections from
different body parts over time and stitching them to form the full human body.
\cite{RF-Pose3D,RF-Pose} also use deep convolutional neural networks to label
various body parts and map them to 3D models of the human skeleton that can be
tracked over time. Our work is different in two aspects. First, unlike humans in 
indoor settings, cars move as one single rigid body and only a single viewpoint
of the car is typically observed in practice. Therefore, even during motion, most portions 
of the car will remain invisible to the system. Second, our goal is to generate higher
resolution images rather than label different body parts in the RF image. 
To this end, we adopt a generative architecture (GAN), which is the state-of-the-art 
in image generation.

\vskip 3pt \noindent
{\bf (b) Millimeter Wave Imaging Systems:} Our work is also related to
traditional mmWave imaging systems that can achieve high
resolution~\cite{Board2017,Sheen2007,Savelyev2010,Ghasr2017,Mamandipoor2015}.
These systems, however, only work in the near field ($< 50 cm$), require huge
amount of bandwidth ($> 10$ GHz) and use very bulky human-sized arrays similar
to airport security scanners~\cite{Board2017}. Other systems can achieve high
resolution at longer distances using focal-plane
arrays~\cite{Meng2018,Ghasr2017,Appleby2007}.  However, these require
integrating optical components like a large focusing lens and a mechanically
steerable raster. Hence, they are bulky and perform poorly on mobile platforms
like self driving cars~\cite{Sheen2007}. Finally, some radar applications use
eigen-space image processing algorithms like MUSIC and ESPRIT to improve
resolution\cite{Hoogeboom2012,Burrows2004}.  However, these techniques cannot deal with 
specularity and artifacts resulting from multipath. They also assume
the signal sources are incoherent which is not the case in mmWave
reflections. 

\begin{figure}[t!]
\centering
\includegraphics[width=\columnwidth]{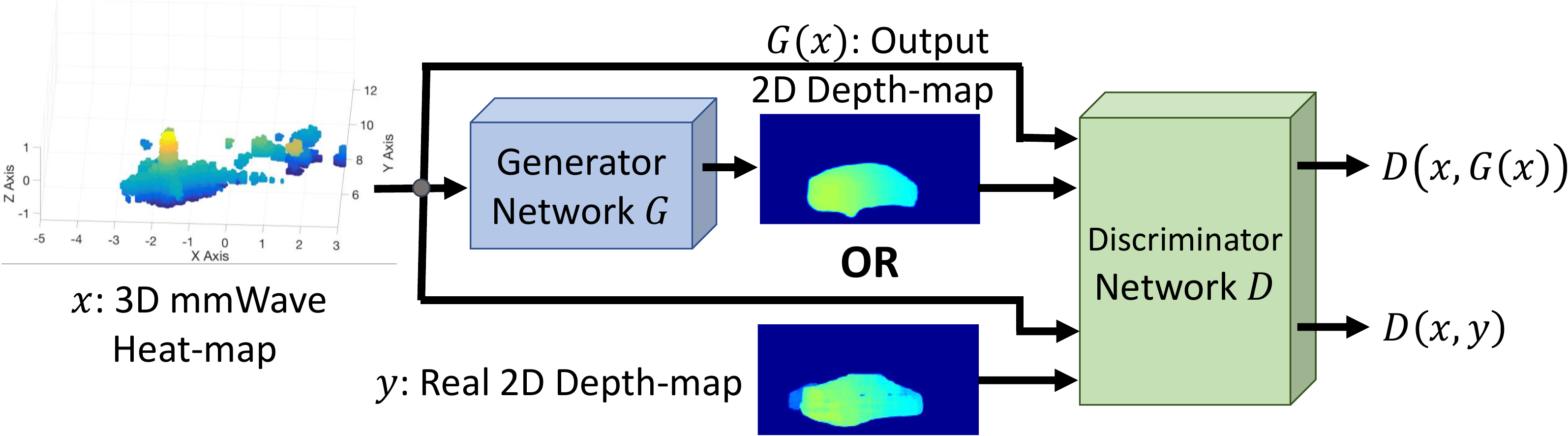}
\vskip -0.1in
\caption{\footnotesize Conditional GAN Architecture.}
\label{fig:cgan}
\vskip -0.2in
\end{figure}

More recent work  uses commercial 60 GHz mmWave radios for imaging
and tracking objects~\cite{Reuse60,60Imaging,mTrack}. However, 
this work is limited to classifying the surface curvature of the object
as concave or convex, localizing the object and tracking it in the
2D plane. { \colorize  The closest to our work are ~\cite{SuperRF, superRF2}. \cite{SuperRF} uses
convolutional neural networks to create  
mmWave SAR images from a smaller SAR array. Similarly, \cite{superRF2} 
applies neural networks to radar acquisitions 
to enhance their resolution. Both \cite{SuperRF, superRF2}, 
however, use radar data both as input and ground-truth to their system,
making them inherently incapable of dealing with challenges 
like specularity and multipath. Moreover, they both focus on 
short-range scenarios.  \name, in contrast, is a mmWave
imaging system that can generate high resolution images
in the far field ($> 10 m$). 
\name's output recovers the visual representation of
the cars and learns to cope with specularity and multipath.}

\vskip 3pt \noindent
{\bf (c) Super-Resolution \& Image Transformation:} In computer vision, neural
networks are used to increase the resolution of camera images and
near-Infrared images~\cite{Ledig2017,Tomoto2016,Glasner2009,Guei2018}.  Such
techniques rely on the correspondence of image patches between low and high
resolution images and can achieve an upscaling factor of $4\times$.
Millimeter wave images, however, have significantly lower spatial resolution
and some of the visual features like boundaries and edges are not apparent.
GANs are also used to transform near-Infrared or low light images to visible
images of the same resolution ~\cite{TV-GAN,Chen2018}. In these applications,
however, the images have the same resolution and share a significant amount of
visual similarities. Finally, in addition to improving resolution, \name\ 
addresses two challenges: missing portions of cars due to specular
reflections and artifacts caused by multipath.

\vskip 3pt \noindent 
{\bf (d) LiDAR in Fog}
There is work on improving the performance of LiDAR in
fog~\cite{Laurenzis2011,Satat2016,Satat}. However, even state-of-the-art
research systems either require knowing a depth map of the scene
apriori~\cite{Laurenzis2011} or work only when the object is static by
estimating the statistical distribution of the photon reflected off the
object~\cite{Satat2016,Satat}. These systems also work only up to 54 cm and have
limited resolution ($32\times 32$ pixels) and field of view. Millimeter
wave, on the other hand, can penetrate through fog, rain, and snow. 
Hence, mmWave radar can augment vision-based systems in self-driving
cars to work in inclement weather.


	\section{Primer on GANs}
\label{sec:primer}


GANs are generative models that have
proven to be very successful since they are able to generate  data samples that
closely follow the distribution of data without explicitly learning the
distribution. To do so, GANs adopt an adversarial learning paradigm 
where two players compete against each other in a {\it minimax} game~\cite{goodfellow2014generative}. 
The first player, the Generator $G$, attempts to generate data samples
that mimic the true distribution, e.g. generate realistic 2D depth-maps of cars. The second player, the Discriminator $D$, 
attempts to differentiate between samples generated by $G$ from
real data samples e.g. differentiate output of $G$ from ground-truth
2D depth-maps from stereo cameras. $G$ and $D$ keep learning until $G$ can generate samples
that $D$ can no longer distinguish from real samples of
the true distribution. At this stage, $D$ is no longer
needed and $G$ can be used during inference to generate
new samples.

\name\ uses a variant called conditional GANs (or cGANs) where $G$ attempts to mimic a
distribution of the data conditioned on some input~\cite{mirza2014conditional}, i.e. $P(y|x)$ where $x$ is
the low resolution mmWave heat-map and $y$ is the high resolution 2D depth-map.
$x$ is given as input to both the generator $G$ and the
discriminator $D$. Fig.~\ref{fig:cgan} shows the architecture of a conditional GAN. 
Intuitively, $D$ serves two purposes. First, it helps generalize 
$G$ by eliminating dependency on the environment i.e., $G$ can create
realistic images of cars by learning features that are 
independent of the background and location of the car in the training scenes. Second, 
it teaches $G$ to fill in the missing parts of the car due to specularity and
eliminate artifacts causes by multipath, i.e., unless artifacts are removed
and specularity is addressed, $D$ will be able to tell that the output
was generated by $G$.  

Mathematically, $G$ learns a mapping from the input 3D mmWave heat-map $x$ to
the output 2D depth-map $G(x)$. $D$, on the other hand, attempts to learn a mapping from 
the input $x$ and a 2D depth-map to a probability $\in [0,1]$ of the 2D depth-map
being real $y$, or generated $G(x)$. A perfect discriminator would give
 $D(x,y) =1$ and $D(x,G(x))=0$. Hence, to win the game against
a given $G$, $D$ tries to maximize the following objective function: 
\[\mathcal{L}(G) = \max_D \Big( \mathbf{E}_{y} \big[\log{D(x,y)}\big] + \mathbf{E}_x \big[\log{(1-D(x,G(x)))}\big]\Big), \]
where the first term is maximized when $D(x,y) = 1$ and the second term is maximized
when $D(x,G(x)) = 0$, i.e.  when $D$ correctly classifies the images as real or generated.

$G$ on the other hand tries to minimize the above objective
function (which is referred to as its loss function $\mathcal{L}(G)$), 
since its goal is to fool the Discriminator into classifying its output data
samples as being real. Therefore, the GAN optimization is a {\it minimax} problem given by:
\[ \min_G \Big( \max_D \Big( \mathbf{E}_{y} \big[\log{D(x,y)}\big] + \mathbf{E}_x \big[\log{(1-D(x,G(x)))}\big]\Big) \Big). \]

Since the mapping functions in $G$ and $D$ can be very complex, $G$ and $D$
are implemented and optimized using deep convolutional neural networks. The
final output of the above GAN optimization is a $G^*$ that minimizes
the loss function $\mathcal{L}(G)$ and can be used to generate 2D depth-maps from new
unseen 3D mmWave heatmaps. Few points are worth noting: 
\begin{itemize}[leftmargin=*,topsep=0pt]
\setlength{\itemsep}{0pt} 
\setlength{\parskip}{0pt}
\item The generator $G$ never actually sees any real ground-truth data.
	Instead, it must learn to create realistic images based only feedback
		it receives from the discriminator $D$. 
\item The GAN never explicitly learns the distribution of the data. Prior to
	GANs, generative models would attempt to explicitly learn the
		distribution by approximating it as mixture of Gaussians or
		other simplified models that typically fail to capture the real
		distribution of data. GANs are very powerful since they
		can generate real looking samples without having to learn
		complex data distributions. 
\item The GAN adaptively learns its own loss function. Any machine learning
	model is trained by optimizing a given loss function that
		provides a quantitative measure of the model's
		performance e.g. $\ell_1$ or $\ell_2$ distance. Choosing the right
		loss function is a very difficult task~\cite{pix2pix}. GANs are powerful
		since they do not require a fixed hand-tuned loss function and 
		rather can adapt the loss function in the above equation as they learn.
\end{itemize}

	\section{System Overview}
\begin{figure}[t!]
\centering
\includegraphics[width=\columnwidth]{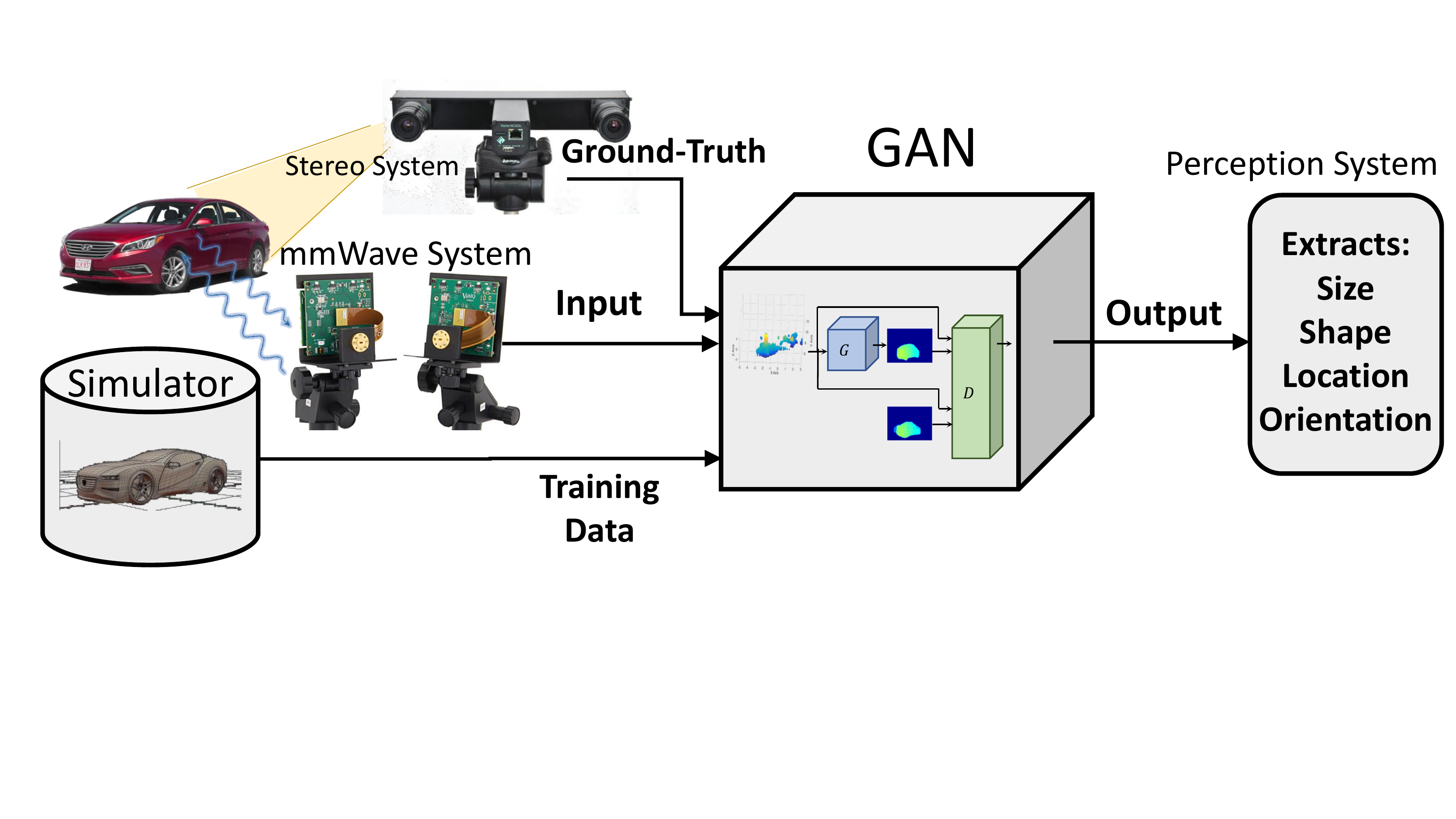}
\vskip -0.7in
\caption{\footnotesize \name's System Overview.}
\label{fig:overview}
\vskip -0.2in
\end{figure}

\name\ is a mmWave imaging system designed for autonomous vehicles. 
\name\ can generate high resolution perceptually interpretable 
2D depth-maps from  3D mmWave heat-maps. To do so, \name\ has
five modules shown in Fig.~\ref{fig:overview}: a mmWave
imaging {\colorize  module}, a stereo camera, a simulator, 
a GAN, and a perception module. The stereo camera and simulator
are used only during training and evaluation. The mmWave,
GAN, and perception modules are used during training and testing. The
perception system is also used for evaluation by extracting
metrics from the ground-truth, input and output. We summarize
these components below:

\begin{itemize}[leftmargin=*,topsep=6pt]
	\setlength{\itemsep}{6pt} 
\setlength{\parskip}{0pt}

\item {\it GAN Architecture} (Sec.~\ref{sec:gan}): We design a new GAN
	architecture customized for mmWave imaging in self-driving
		cars. The GAN uses an encoder-decoder paradigm, a modified
		loss function and skip connection to produce perceptually
		interpretable and accurate reconstructions of cars. 

\item {\it mmWave Imaging {\colorize  Module}} (Sec.~\ref{sec:mmW}): We custom-build
	a mmWave imaging {\colorize  module} using off-the-shelf 60 GHz radio and
		RF circuit components. It also uses a linear slider platform
		to emulate a large antenna array leveraging synthetic aperture
		radar, which provides us with more flexibility to test
		and experiment various parameters and setups. {\colorize This module} produces
		3D mmWave heat-maps. However, on its own, it is intrinsically limited
		in resolution and suffers from specularity and artifacts
		caused by multipath.

\item {\it Simulator} (Sec.~\ref{sec:sim}): This module augments the training
	dataset with synthesized data obtained from 3D CAD models of cars and 
		mmWave ray tracing algorithms. It produces both the ground truth
		2D depth-map and the synthesized 3D mmWave depth-maps.

\item {\it Stereo Camera} (Sec.~\ref{sec:imp}): \colorize  We custom-build a long-range
	stereo camera that can generate high-resolution 2D depth-maps
		of cars that serve as ground-truth for \name. \color{black}

\item {\it Perception Module} (Sec.~\ref{sec:quant}): This module
	allows us to extract quantitative metrics such as
	size, shape, location, orientation and boundary of the car. 
	We use it to evaluate  {\colorize the output of \name\ by comparing the output
with the ground truth and the input mmWave 3D heat-maps. }

\end{itemize}
Next, we will discuss these modules in more detail.

	\section{\name's  GAN Architecture}
\label{sec:gan}

\begin{figure*}[t!]
	\centering
	\includegraphics[width=7in]{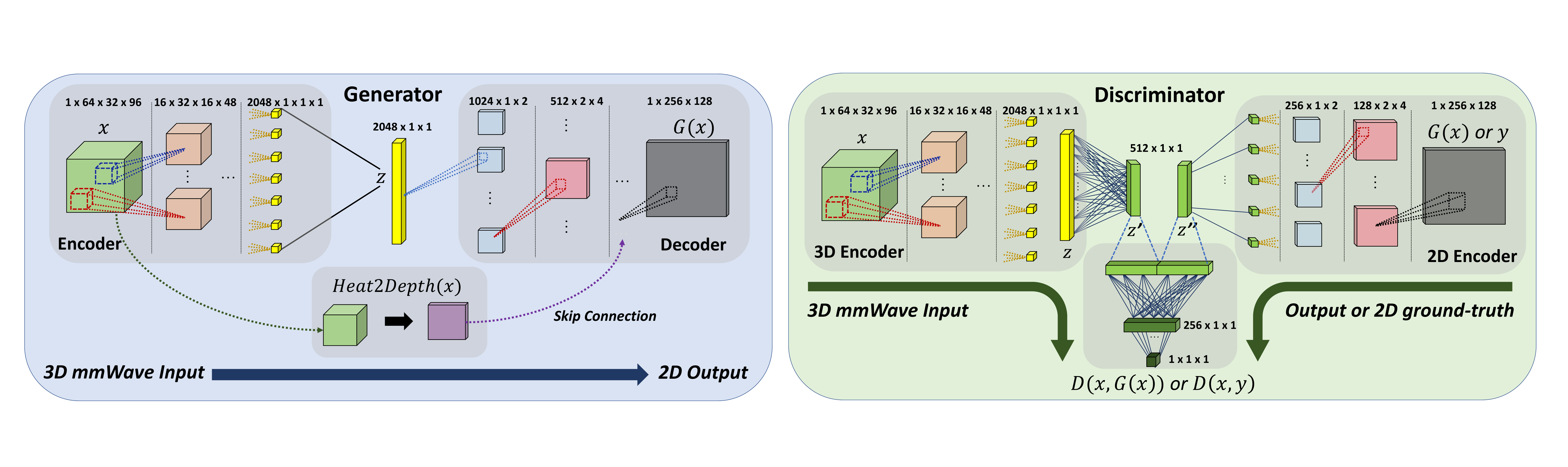}
	\vskip -0.1in
	\caption{\footnotesize Network Architecture of the Generator $G$ and Discriminator $D$.}
	\label{fig:g}
	\vskip -0.1in
\end{figure*}

\name\ leverages a cGAN architecture to enhance the resolution and quality of mmWave imaging
in self-driving cars. However, unlike image-to-image mapping in computer vision, mmWave RF 
imaging brings in new challenges such as perspective mismatch between input and
output, along with quantitative accuracy as described earlier.
To address these challenges, \name\ introduces a new GAN
 architecture tailored for mmWave imaging in self-driving cars.

\name's GAN must learn to generate a high-resolution 2D depth-map from a 3D mmWave heat-map. 
That is, \name's ground-truth and output are 2D images with each pixel 
representing depth, whereas its input is a 3D image where each voxel maps to RF signal strength.
However, cGAN architectures typically map 2D input to 2D
output~\cite{yoo2016pixel,zhu2017toward,zhang2017image} or 3D input to 3D output~\cite{jin2018ct}. 
One solution to address the discrepancy in dimensions would be to
transform the 2D depth-maps to 3D point clouds where each voxel value maps to a
binary one or zero and use a point cloud GAN similar to~\cite{PointGAN} to
generate 3D point clouds from 3D heat-maps.

Unfortunately, such design results in sparse high-dimensional input and output
data. Training such 3D GAN is known to be notoriously hard~\cite{smith2017improved}. As the
sizes of the input and output increase, the number of parameters in the neural network of the generator $G$ and discriminator
$D$ increase significantly. Consequently, the learning process that optimizes
$G$ and $D$ must search through an exponentially larger space. Hence, 
it is more difficult for the network to converge and heuristics like gradient 
descent become more prone to getting stuck in local minima. For example, 
our input 3D heat-map $x$ has a size of $64\times 32 \times 96$.  
Increasing the resolution of the heat-map by $4\times$ would output
a 3D point cloud $G(z)$ of size $256\times 128 \times 384$. Training the neural network
underlying $G$ with this input and output size would require $\approx 10$ billion parameters
making the training intractable.


{\colorize Instead, \name\ first separately encodes both the 2D input and 3D output into a common 1D space through separate neural networks (Fig.~\ref{fig:g}). 
This design has two benefits. First, it significantly reduces the complexity of 
the network~\cite{badrinarayanan2017segnet}, and second, it allows
for the encoded input and output to be compared directly. The resulting networks require $\approx 260$ million 
parameters and can therefore be run on a standard GPU. Details of network design follow recommendations proposed in DCGAN~\cite{dcgan}. }

\vskip 6pt \noindent{\bf A. Generator Architecture:}

{\colorize As described in section~\ref{sec:primer}, the generator is implemented using deep neural networks.
Following the previous paragraph, instead of directly generating the output from the input $x$
using consecutive layers, the input is first encoded into a low-dimension
representation $z$ using an encoder network. $z$ is then decoded into the output
$G(x)$ using a decoder network.}  {\colorize This type of structure is also known as the encoder-decoder architecture~\cite{badrinarayanan2017segnet}}.

The vector $z$ represents a common feature space
between input and output. The input $x$ represents the 3D mmWave heat-map in spherical coordinates (azimuth angle,
elevation angle, range) generated by the mmWave module, where voxel values represent the RF signal
strength as we describe in more detail in section~\ref{sec:mmW}. The output
$G(x)$ represents a higher resolution 2D depth-map where pixel values represent depth.

The encoder starts with one channel of 3D input. {\colorize  At each layer, there are 3D convolutions each of which is followed by 
Leaky-ReLU activation functions and batch-norm layers. With each layer, the number of channels increases and the size of the 3D kernel decreases until
 we get to an $n \times 1\times 1 \times 1$ ($n=2048$) vector $z$ as shown in Fig.~\ref{fig:g}.}
$z$ is then squeezed to an $n \times 1 \times 1$ vector to account for the dimension
mismatch and passed to the decoder where it goes through the reverse
process albeit using 2D transposed convolutions. {\colorize Each layer in the decoder increases
the size of the 2D kernel and decreases the number of channels until the output $G(x)$. 
In particular, the generator uses 5
convolutional layers in the encoder and 7 transposed convolutional layers in the decoder.}

{\colorize  \name's generator also adapts the idea of \emph{skip connections} between the
encoder and the decoder~\cite{ronneberger2015u}. Specifically, \emph{skip connections}
allow for information in an early layer to fast-forward directly to output layers.
In our case, mmWave provides very accurate range information which, 
ideally, should be mapped directly to the output. As going through consecutive layers tends
to saturate the accuracy of the approximation~\cite{he2016deep}, we add a skip
connection from input of the encoder to third-to-last layer of decoder. }

Skip connections are generally implemented by simply
concatenating output feature map of one layer to the input feature
map of another layer of the same dimensions. However, in \name's
architecture, none of the layers of the encoder and decoder have
same dimension due to the dimension mismatch between the 3D
input and 2D output. To address this, \name\ introduces a 
heat-map to depth-map transformation. {\colorize Specifically, for
each 2D spatial location in the input 3D heat-map, we
assign the depth as the location of the largest value (corresponding to highest signal strength)
along the depth dimension.} Formally, 
$$
	x_{2D}(\phi, \theta) = \arg\max_r\ x_{3D}(\phi,\theta, r).	
$$
However, simply choosing the depth corresponding to 
the largest value is unstable and can lead to errors. Instead, 
we chose the $m$ largest values and create $m$ channels of
$2D$ depth-maps which we can then concatenate to a feature map
of the same dimension in the decoder. 
It is worth noting that the above transformation only
makes sense when applied to the input. We choose $n=8$ in our implementation.


\vskip 12pt \noindent{\bf B. Discriminator Architecture:}

The discriminator takes two inputs: the 3D mmWave
heat-map $x$ and a 2D depth-map that either comes from the ground-truth $y$ 
or was generated $G(x)$. It outputs a probability of the input being real. 
In standard cGANs, the input and output of $G$ typically have
the same dimension. Hence, they are simply concatenated and fed as one
input to $D$ that uses a single deep neural network to output $D(x,y)$ or
$D(x,G(x))$. 

However, \name's input and output have different dimensions and hence cannot
simply be concatenated and mixed together by one neural network. {\colorize  Instead,
\name\ uses two encoder networks inside the discriminator $D$ as shown
in Fig.~\ref{fig:g}.  The first is a 3D encoder with the same architecture as that in the Generator. It outputs a low-dimension representation $z$. The second
is a 2D encoder with 7 convolutional layers that takes the 2D depth-map $y$ or $G(x)$ as input and outputs
a low-dimension representation $z''$.} Before mixing the output of the two
encoders, \name\ ensures that they map to the same feature space by converting $z$ to $z'$ using a fully connected layer as shown in
Fig.~\ref{fig:g}. $z'$ and $z''$ are then concatenated and mapped to 
the output probability of the discriminator using two fully connected layers.

\vskip 6pt \noindent{\bf C. Loss Function:}

The output of the discriminator $D$ and generator $G$ are used to calculate the
loss function $\mathcal{L}(G)$ defined in section~\ref{sec:primer}.  During
training, $D$ and $G$ are optimized to minimize this loss function.  As
mentioned earlier, GANs are very powerful at creating realistic images by
capturing the high frequency components of the scene.  However, they tend to be
less accurate when it comes to low frequency components such as coarse object
location, orientation, etc. The reason for this is that 
the objective of the generator $G$ is to simply create high resolution images
that look \emph{real} to the discriminator $D$ rather than accurate images
with respect to metrics like location, orientation, etc. {\colorize  To address this, past work in machine learning manually adds an $\mathcal{L}_1$ loss
term or a perceptual loss term $\mathcal{L}_p$~\cite{pix2pix,johnson2016perceptual}. $\mathcal{L}_1$ loss is defined as the
$\ell_1$ distance between the ground truth and the output of the GAN (Eq. \ref{eq:l1_loss}).
\begin{align}
& \mathcal{L}_1(G) = \mathbf{E}\|y - G(x) \|_1  \label{eq:l1_loss}\\
& \mathcal{L}_p(G) = \mathbf{E}\|VGG(y) - VGG(G(x)) \|_1 \ \label{eq:vgg_loss} \\
& \mathcal{L}_H(G) = \mathcal{L}(G) + \lambda_1 \mathcal{L}_1 + \lambda_p \mathcal{L}_p \label{eq:g_loss} 
\end{align}

For perceptual loss, a standard practice is to use a pre-trained network (e.g. VGG network
from~\cite{simonyan2014very}) that extracts perceptual features of the image.} The ground truth
and output are passed through this network and $\mathcal{L}_p$ is
computed as the $\ell_1$ distance between the feature maps that the network outputs (Eq \ref{eq:vgg_loss}).
The recent trend in computer vision is to avoid directly using $\mathcal{L}_1$ loss
and instead use perceptual loss~\cite{johnson2016perceptual,zhang2018unreasonable}. This is because the difference
between individual pixel values in images carries little to no perceptual information. 
{\colorize  Unlike images, pixel values in depth-maps correspond to depth and carry perceptual information about
cars, such as orientation and shape.} Hence, \name\ maintains a combination of three losses (Eq. \ref{eq:g_loss}), where $\lambda_1$ and $\lambda_p$ are hand-tuned relative weights of the loss functions. 
Using this loss function enables \name\ to accurately capture both the low and high frequency
components in the image. This results in perceptually interpretable high resolution images that
faithfully represent the scene.

	\section{mmWave Imaging Module}
\label{sec:mmW}
As described earlier, the advent of 5G in the mmWave spectrum 
has led to creation of electronically steerable phased arrays with hundreds
of antenna elements~\cite{Bell,UCSD}. { \colorize  The very short wavelength of mmWave signals allows these phased
arrays to have very small form factor. For example, 
at 60 GHz, a $32\times 32$ array
only occupies an $8cm\times 8cm$ patch.}

\vskip 6pt \noindent{\bf Imaging:}
{ \colorize The ability to create very narrow beams and steer them electronically enables
mmWave phased array radios to image 3D objects in the environment.} Specifically, antenna
array theory tells us that for $N\times N$ array, we can compute the 
reflected power along the spherical angles $\theta$ (elevation) and $\phi$ (azimuth) by
adding a phase shift to the signal received on every antenna before
combining the signals~\cite{ArrayDSP}. Formally, 
\[x(\theta, \phi) = \sum_k^N\sum_l^N{S_{k,l}e^{j\frac{2\pi}{\lambda} [(k-1) d
sin(\theta)cos(\phi) + (l-1) d cos(\theta)]}}\]
where $\lambda$ is the wavelength, $d =\lambda/2$ is the separation
between consecutive elements, and $S_{k,l}$ is the signal received on 
the antenna element index by $(k,l)$ in the 2D array.

The third dimension (the range $\rho$) is obtained by measuring the time
of flight of the radar waveform echo. The huge bandwidth available in the mmWave band allows us to estimate range with high resolution. In our design we transmit a low power FMCW (Frequency Modulated
Continuous Wave) similar to the one in~\cite{WiTrack}, albeit at 60 GHz. We adopt a heterodyne architecture where { \colorize we first generate the FMCW waveform at the baseband and then
up-convert it to mmWave frequencies. The received signal is down-converted to baseband for radar waveform processing. This allows us to easily change
the frequency band of operation to 24 GHz or 77 GHz by changing
the mmWave front-end.} The architecture is described in more details in 
section~\ref{sec:imp}.  { \colorize The time of flight can be extracted from the
FMCW using a simple FFT on the beat signal sampled below 1 MHz as described in~\cite{WiTrack}.  
This allows us to reconstruct a 3D heat-map $x(\theta, \phi, \rho)$.}

\vskip 6pt \noindent{\bf Challenges:} As described earlier, 
mmWave imaging suffers from 3 challenges. The first is resolution. 
The angular resolution is set by the size (aperture) of the antenna array and the range resolution is set by the bandwidth of the radar waveform. Specifically, for a bandwidth $B$, the
range resolution is given by $\frac{c}{2B}$ where $c$ is the speed 
of light. The range resolution for our system is 10 cm (with $B=1.5$ GHz) which is $3.3\times$ worse
than that of the commercial LiDAR~\cite{Velodyne}.  

\begin{figure}[t!]
	\centering
	\includegraphics[width=3in]{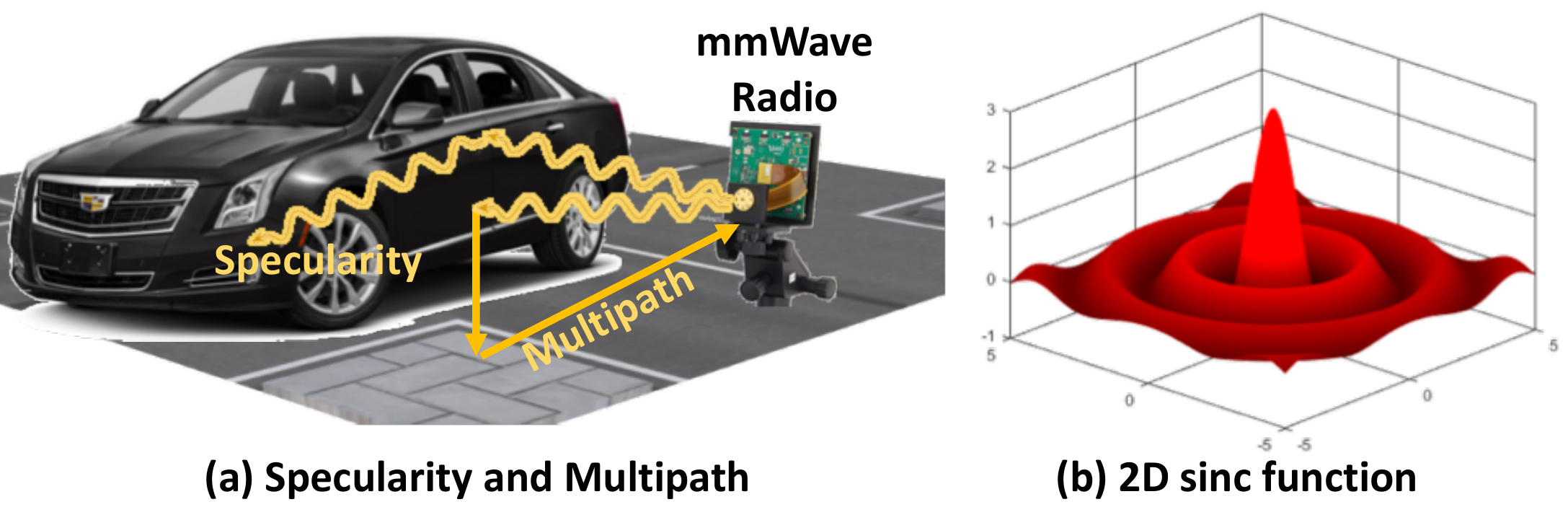}
	\vskip -0.1in
	\caption{\footnotesize Challenges in mmWave Imaging}
	\label{fig:mmWC}
	\vskip -0.2in
\end{figure}

Resolution along the other two dimensions, however, is significantly worse.
The Rayleigh limit of the angular resolution is given by $\pi \lambda/L$ where
$L$ is the antenna aperture given by $L = N \lambda/2$ \cite{ArrayDSP}.
{ \colorize  Even with the small antenna array form factor at mmWave frequency,
 we would need a 9 m long antenna array to achieve sub-degree angular resolution similar to LiDAR~\cite{Velodyne}.}
\footnote{Note that for systems like the airport security scanners, 
the target being imaged is in short range and hence, human sized arrays
are sufficient.} Today's practical mmWave systems have a much smaller (around $100\times$) aperture
which limits the resolution, resulting in the image getting convolved
with a very wide 2D sinc function similar to the one shown in Fig.~\ref{fig:mmWC}~(b). That is
why the mmWave image along the azimuth and elevation dimension mainly looks like
blobs as shown in Fig.~\ref{fig:teaser}~(c) where most of the perceptual content
such as object boundaries are lost.

Resolution, however, is not the only challenge. Unlike light, mmWave signals do
not scatter as much and mainly reflect off surfaces. This leads to specularity as shown in
Fig.~\ref{fig:mmWC}~(a) where some reflections never trace back to the mmWave receiver,
making certain portions of the car impossible to image. Moreover, 
due to multipath propagation, some reflections bounce off the street and other obstacles
and trace back to the receiver as shown in Fig.~\ref{fig:mmWC}~(a) creating many 
artifacts in the mmWave image as shown in Fig.~\ref{fig:teaser}~(c). 

\vskip 6pt \noindent{\bf Synthetic Aperture Radar (SAR):} Due to the limited availability of large 2D phased arrays in commercial systems, \name\ emulates it using the synthetic aperture radar approach.
{ \colorize  As opposed to the analog beam-forming in phased array radios, we implement digital beam-forming at the receiver. We synthesize a large antenna array with a 60 GHz front-end and an FMCW baseband circuit, by physically scanning a single antenna element. The signal from each antenna element location is sampled, and phase-shifts are applied in digital afterwards.}
While this presents a limitation that prevents our implementation from operating in non-static scenarios, 
it gives us the flexibility to experiment with various array sizes. We
discuss the implementation of SAR and this limitation in sections~\ref{sec:imp} and~\ref{sec:limit} respectively.


	\section{\name's Data Simulator}
\label{sec:sim}
As mentioned earlier, the amount of data required to train our GAN is in
the order of thousands of input and groundtruth images. Collecting
such huge amount of data would take a prohibitively long time. 
To address this, we build a simulator that synthesizes paired up 3D mmWave heat-maps
and 2D depth-maps of cars from 3D CAD models.

Our simulator is designed to create 3D point reflector
models of cars and then simulate reflected
mmWave signals using ray tracing. It takes into account
multipath reflections as well as specularity based on reflection angles
 to generate realistic mmWave 3D heat-maps. The simulation has 3 stages: 
\begin{itemize}[leftmargin=*,topsep=3pt]
	\setlength{\itemsep}{0pt} 
	\setlength{\parskip}{0pt}

\item[(1)]{\it Scene generation}: We first simulate scenes of cars
based on two types of datasets: 3D CAD model for an autonomous driving 
dataset~\cite{Geiger12CVPR} and Cityscapes~\cite{cityscapes}, a street view video recordings dataset.
 The 3D CAD models provide us with precise 3D meshes
of a wide variety of vehicles, while the street view photos offer references
for car placement through object masks captured with Mask R-CNN~\cite{maskRCNN}.


\item[(2)]{\it Ray Tracing}: Here we model the radar cross-sections of the scene.
Occluded bodies are removed through spherical projection. Then, 
the radar cross-sections of the remaining surfaces are modeled as a cluster of
point reflectors with different densities and reflectivities. 
{ \colorize We pinpoint corner areas with more scattering and specular
surface areas whose reflection will not be received, and model point
reflectors to perform standard ray tracing~\cite{mmWaveRayTracing}.}

\item[(3)]{\it Ground-truth \& mmWave image generation}:
{ \colorize We then simulate the received signal based on the point reflector model with background noise introduced. We add thermal noise to the FMCW signals and error in antenna element positions to better match the real data. By applying standard mmWave image processing as described in section~\ref{sec:mmW} we get the 3D mmWave heat-map.} The ground-truth 2D depth-map is generated through spherical projection of the 3D scene and coloring the pixels according to the depth.

\end{itemize}

	\begin{figure*}[t!]
\centering
	\begin{minipage}{4.6in}
	
\includegraphics[width=4.6in]{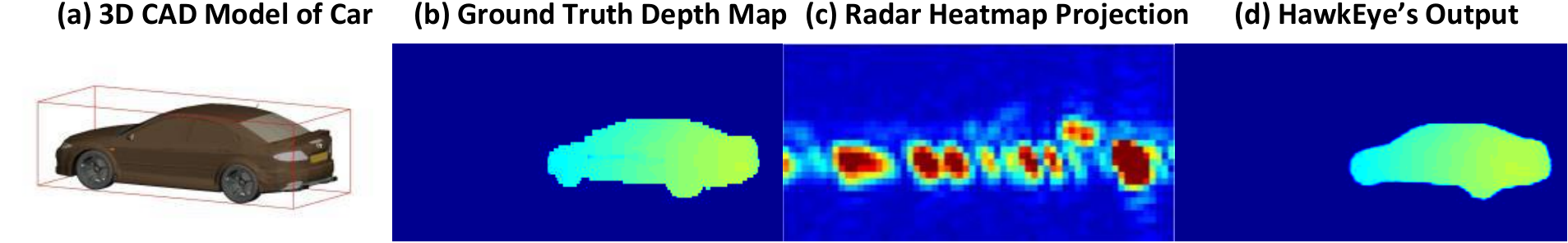}
\vskip -0.15in
\caption{\footnotesize \name's qualitative performance on synthetic test data. }
\label{fig:qual_synth}
\vskip 0.05in	
\includegraphics[width=4.6in]{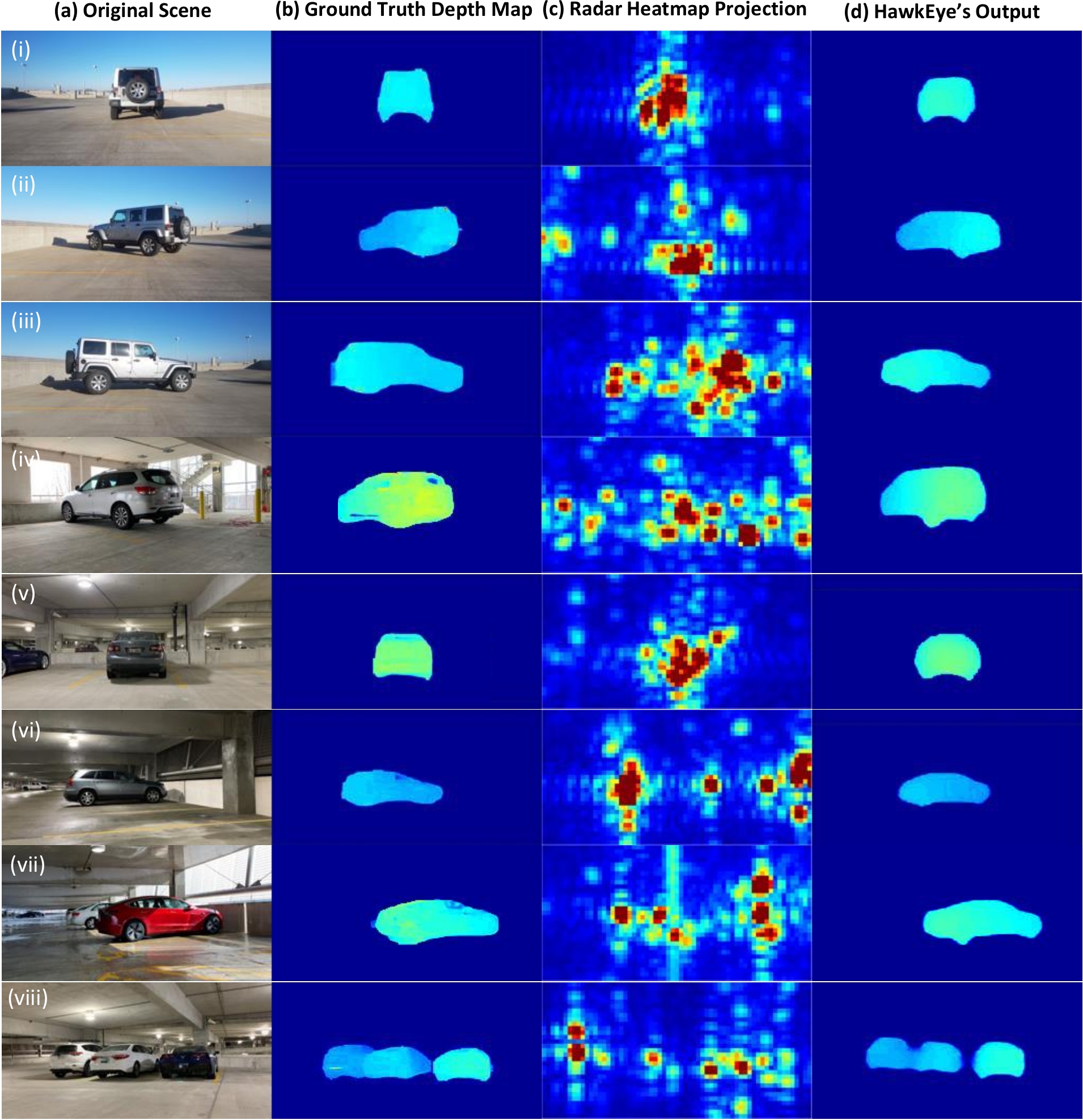}
\vskip -0.15in
\caption{\footnotesize \name's qualitative performance on real test data. }
\label{fig:qual_real}
\vskip -0.1in
	\end{minipage}	
\begin{minipage}{2.2in}
\centering

\includegraphics[width=1.7in]{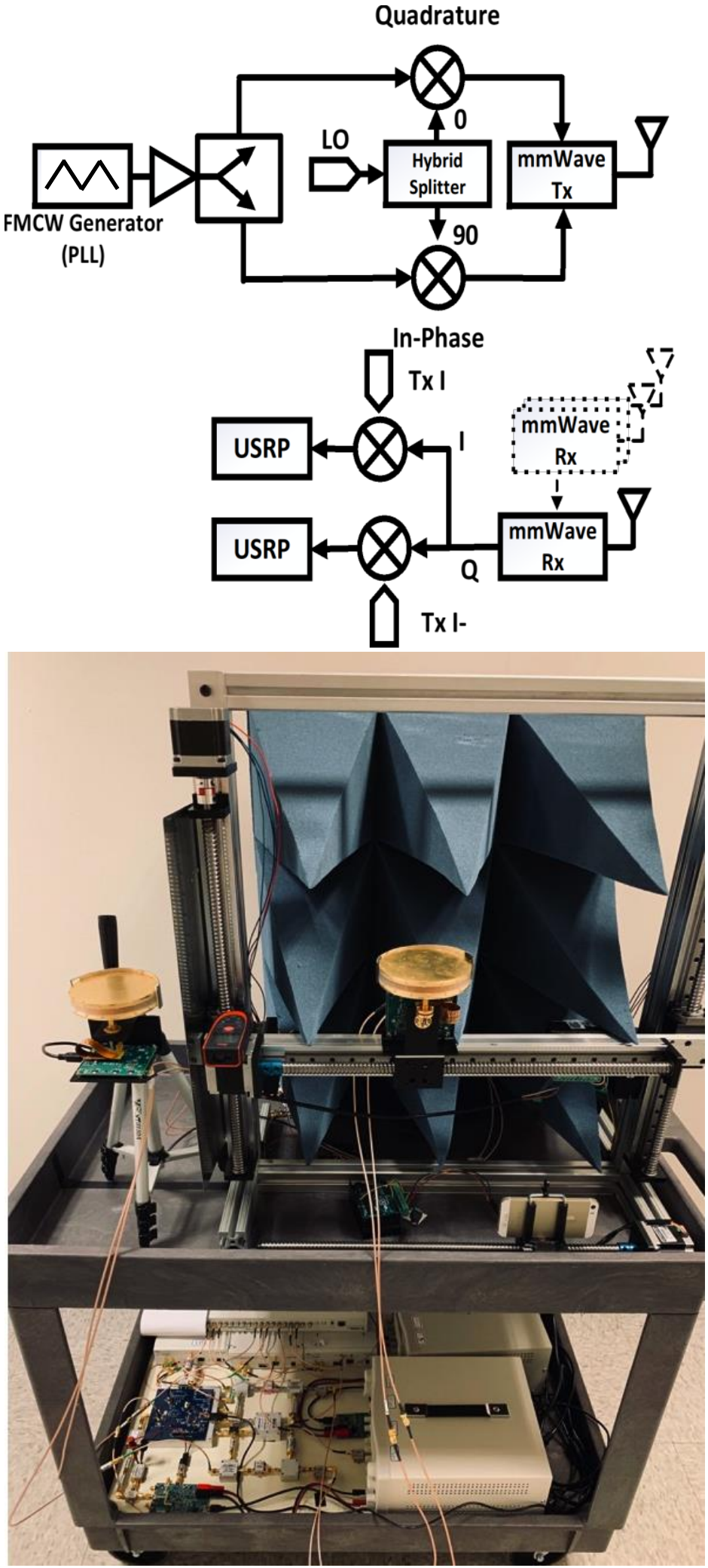}
\vskip -0.1in
\caption{\footnotesize Millimeter Wave Platform}
\label{fig:SAR_frame}
\vskip 0.05in

\includegraphics[width=1.8in]{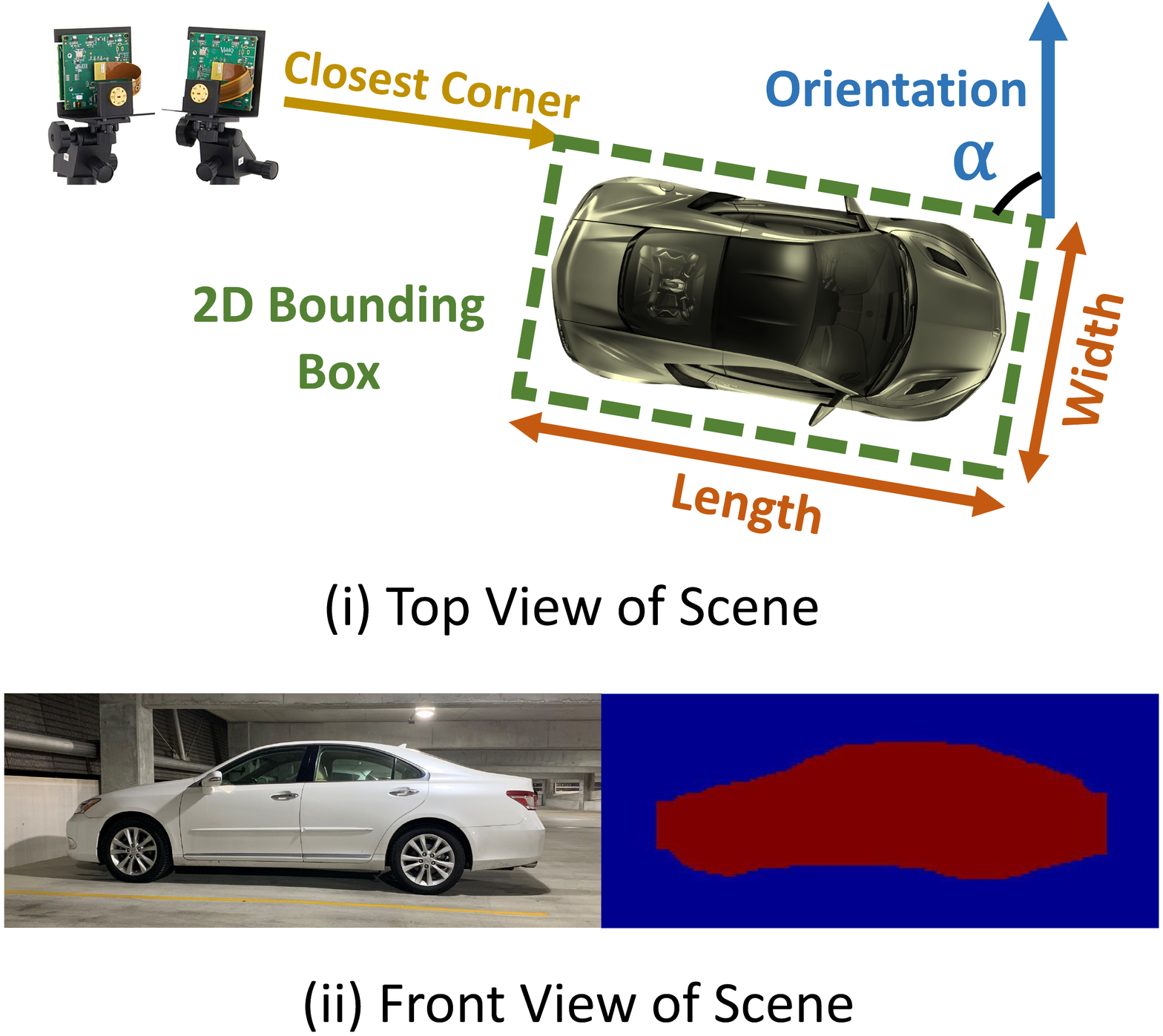}
\vskip -0.15in
\caption{\footnotesize Metrics}
\label{fig:metrics}
\vskip -0.1in
\end{minipage}
\vskip -0.15in
\end{figure*}

\section{Implementation}
\label{sec:imp}

\noindent{\bf A. GAN Implementation \& Training:}
The output of the discriminator and generator are used to calculate loss
function $\mathcal{L}(H)$ as described in section~\ref{sec:primer}. The
generator and discriminator are trained end-to-end by optimizing the loss
function. We follow the standard back-propagation algorithm and train on an Nvidia Titan RTX GPU.

{ \colorize The training dataset is formed of 3000 synthesized images of cars with a batch size of 4.} We use 120 different
car models while varying the orientation and location of each car. 
After 170 epochs, we fine tune the GAN using 100 real mmWave images. We test on 500 synthesized
images and 140 real images. In the test methodology, we follow standard k-fold cross-validation where $k=5$. We ensure that the examples in the test dataset are not used during training. 
The real images come from 40 different car models with various orientations and locations.

\vskip 6pt \noindent{\bf B. mmWave Imaging:}
 As described earlier the mmWave imaging radar comprises 60 GHz radios and SAR. For SAR, we build a 2D linear slider platform 
 shown in Fig.~\ref{fig:SAR_frame} using three FUYU FSL40 linear
 sliders with sub-millimeter accuracy. A horizontal slider scans the mounted
 receiver antenna along the X-axis, while two vertical sliders scan along the Z-axis. 
 {\colorize  This platform allows us to synthesize a reconfigurable antenna array with a wide range of frequency and aperture. In \name, only a fraction of $20 cm \times 20 cm$ area is scanned to emulate a $40\times 40$ array at 60 GHz. The scanning time is 5 minutes, and will reduce to 90 seconds for a $20\times 20$ array.} 

\begin{figure*}[t!]
\centering
	\begin{minipage}{4.6in}
	\includegraphics[width=4.4in]{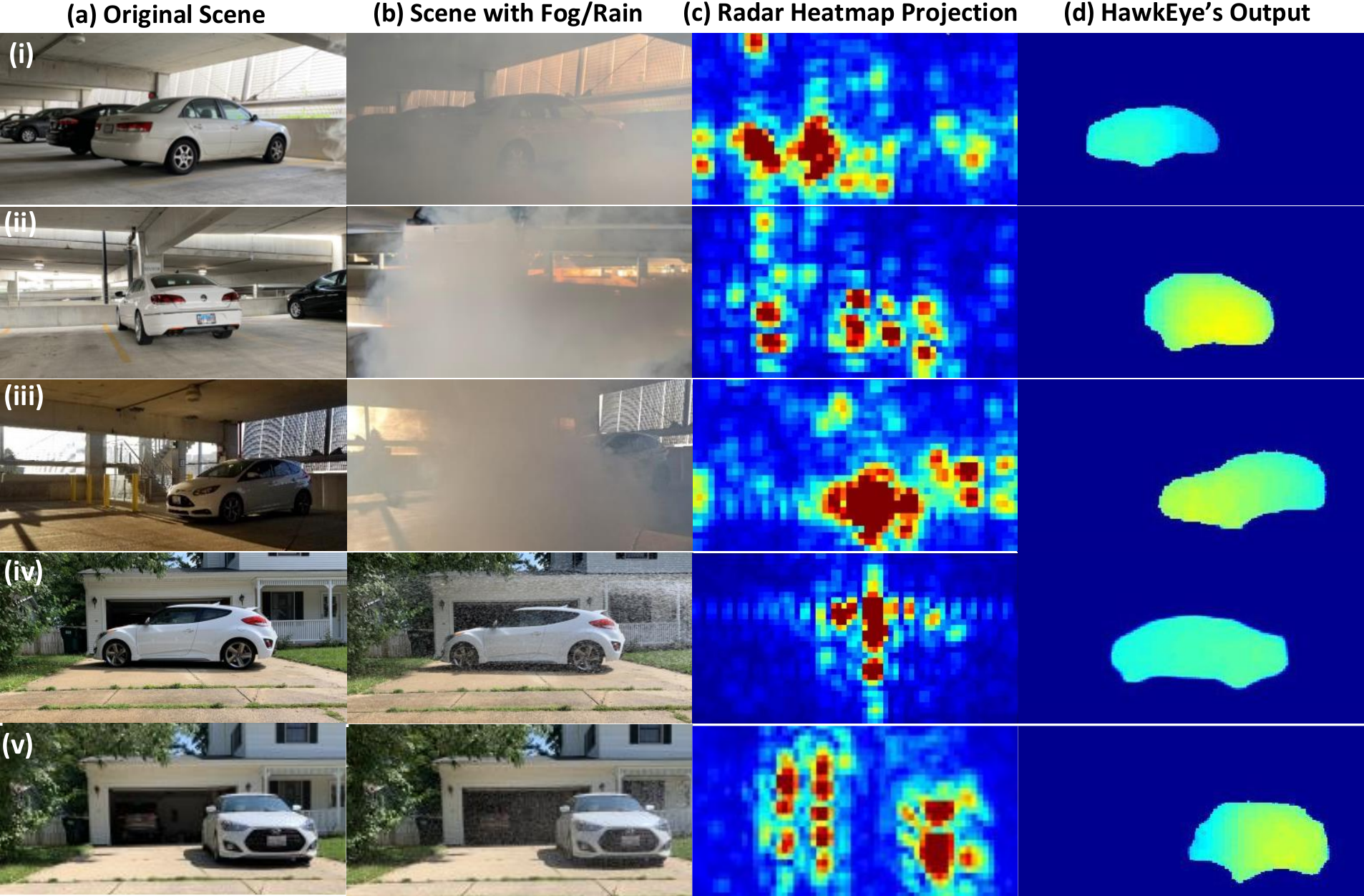}
	\vskip -0.05in
	\caption{\footnotesize \name's performance with fog/rain in scene. }
	\label{fig:qual_fog}
	\vskip -0.1in
	\end{minipage}
	\begin{minipage}{2.2in}	
\centering
	\includegraphics[width=2.1in]{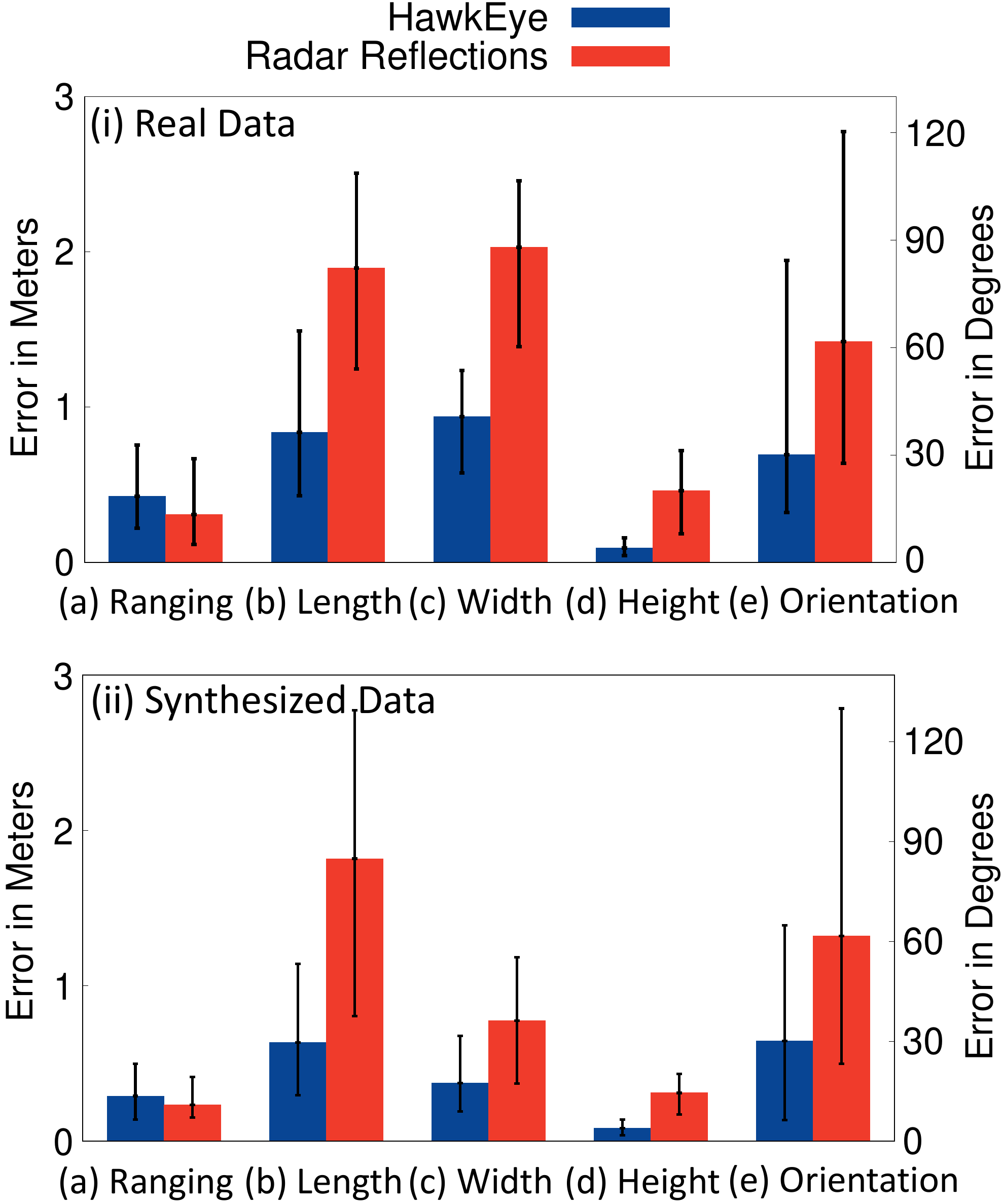}
	\vskip -0.13in
	\caption{\footnotesize Quantitative Results.}
	\label{fig:bar}
	\vskip -0.1in
	\end{minipage}
	\vskip -0.15in
\end{figure*}

{ \colorize  For the mmWave circuit, we implement a heterodyne architecture as shown in Fig.~\ref{fig:SAR_frame}. 
We first generate the FMCW with triangular waveform at baseband, sweeping a bandwidth of 1.5 GHz. Then we up-convert it to 60 GHz with quadrature modulation using Pasternack 60 GHz radio frontend. The resulting signal after sideband suppression sweeps from 59.5 GHz to 61 GHz. In the receiver, 
we implement quadrature demodulation and sample the I and Q components of the complex signal with two synchronized USRPs for direct phase measurement and shifting.}

\vskip 6pt \noindent{\bf C. Stereo Camera \& Generating Ground-truth: }
Ideally, we would use LiDAR to generate the ground-truth. However, LiDAR
systems are extremely expensive and avoided by many autonomous car
manufactures like Tesla. Hence, we opt for a stereo camera system to generate
2D depth-maps for our ground-truth. We place an iPhone X on a linear slider to capture multiple images of the scene. We then apply standard stereo image processing to extract the 2D depth-map. 
{\colorize We filter out pixels that do not belong to the vehicles of interest using labeled object masks, that we create by applying a pre-trained object detection model, Mask R-CNN~\cite{maskRCNN} on raw camera images.}

It is important to note that the mmWave imaging module and the stereo
camera module, even when placed at the same location, will not yield the same
view point and field of view. Hence, we calibrate for such discrepancy to be
able to accurately train and test the GAN.

\section{Evaluation}

We compare \name\ against the ground truth obtained from the stereo camera system as well as images generated by
the mmWave imaging system. {\it  All evaluation is performed on outputs from the testing dataset that were 
not shown to the GAN during training.} We present both qualitative and quantitative results. The quantitative
results are extracted using the perception module described in section~\ref{sec:quant}. 

\subsection{Qualitative Results}
\label{sec:qual}

We show \name's performance on the
synthetic test data generated from our simulator in Fig.~\ref{fig:qual_synth}
and on real radar heat-maps captured in outdoor environments in
Fig.~\ref{fig:qual_real}. Column (a) shows the car being imaged (the license plates are blurred), and Column (b) shows the corresponding
2D \colorize  stereo \color{black} depth-map of the car. Columns (c) and (d) show the 2D \colorize  front-view \color{black} projection of the radar
heat-map and the output of \name\, respectively. We can see that \name\
accurately reconstructs the shape and size of the car in the scene, and captures key defining features such as its wheels and
orientation. \name\ can also accurately determine the distance of the car in
3D space, as can be seen from the depth-maps.

\begin{figure*}[t!]
\centering
	\begin{minipage}{4.7in}
	\includegraphics[width=4.4in]{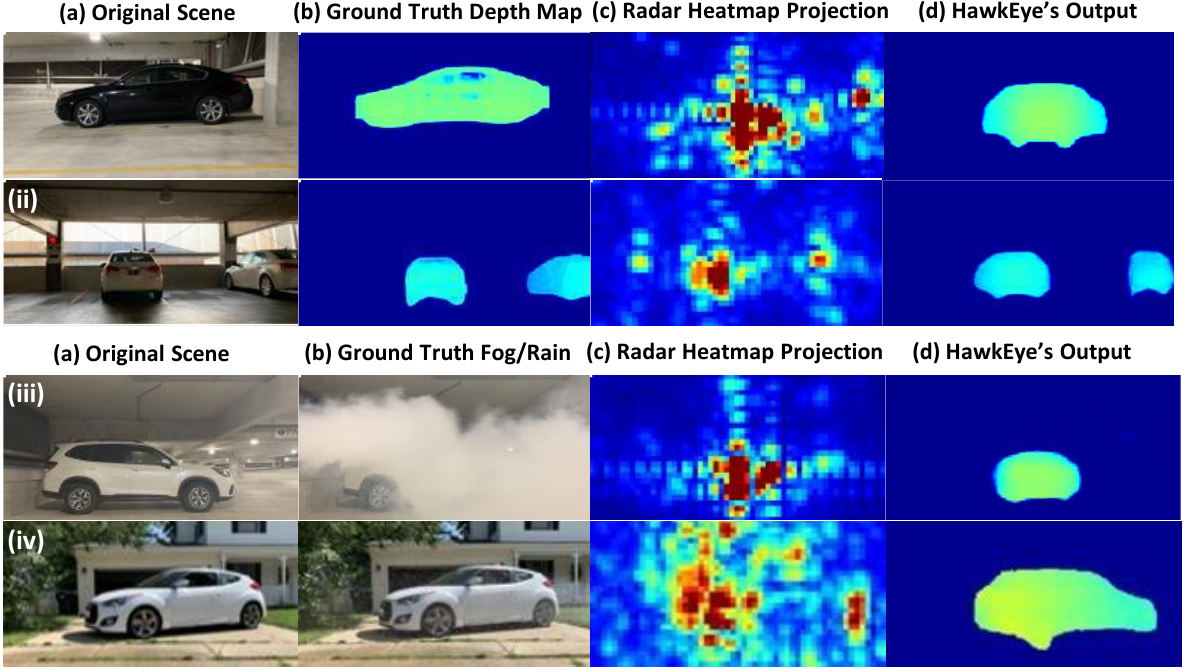}
	\vskip -0.1in
	\caption{\footnotesize Examples where \name\ fails. }
	\label{fig:failure}
	\vskip -0.1in
	\end{minipage}
	\begin{minipage}{2.0in}	
	\centering
			\includegraphics[width=1.8in]{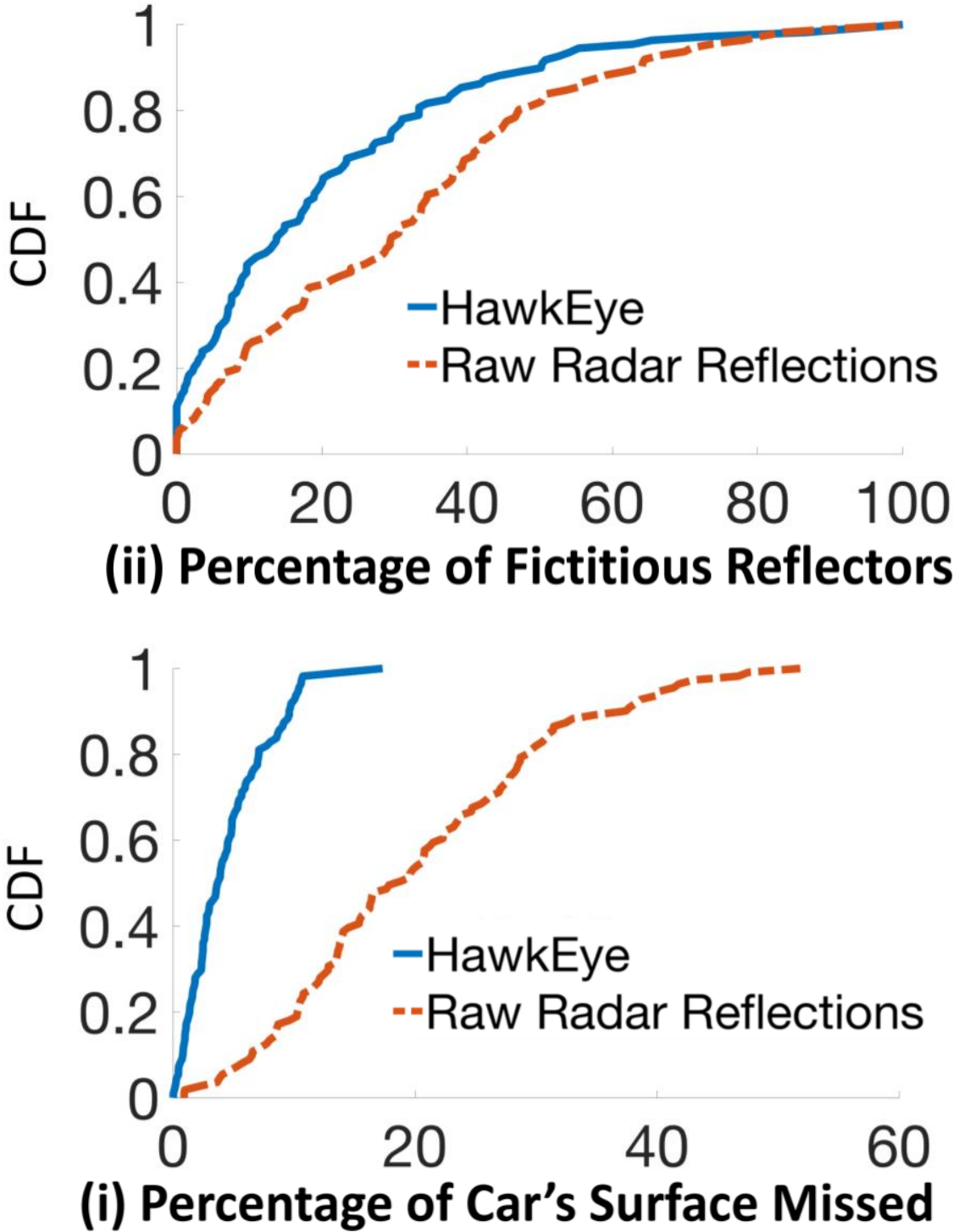}
		\vskip -0.12in
		\caption{\footnotesize Accuracy of shape of car.}
		\label{fig:shape}
		\vskip -0.1in

	\end{minipage}
	\vskip -0.15in
\end{figure*}

These results show that even though the GAN was trained primarily on synthesized 
training data, it could generalize well to real data with only a small amount
of fine-tuning during training. Hence, the simulator
faithfully emulates the propagation characteristics of mmWave signals by using
precise ray tracing models and incorporating other artifacts such as
specularity and multipath reflections. 

\colorize 
\vskip 4 pt
\noindent
{\bf Results in Fog/Rain:} We also conduct experiments in fog and rain to test \name's performance in poor visibility conditions where today's optical sensors fail. Due to practical limitations and to avoid the risk of water damage to our setup, we conduct controlled experiments where we emulate real fog and rain. 
We use a fog machine along with a high-density water-based fog fluid to emulate severe and realistic fog conditions. We emulate rain using a water hose in a confined region around the object of interest (the car).

It is important to note that \name's GAN model was never trained with examples collected in fog or rain. During training, \name\ only uses synthetic data and is fine-tuned with real data collected in clear weather conditions.We present experimental results for fog (Fig.\ref{fig:qual_fog} (i)-(iii)) and rain (Fig.\ref{fig:qual_fog} (iv)-(v)). Column (a) and (b) show the original clear scene with the car, and the same scene with fog/rain respectively. Column (c) shows the 2D projection of the mmWave radar heat-map captured in the scene with fog/rain, and Column (d) shows the corresponding output of \name. From Fig.\ref{fig:qual_fog}, we can see that \name\ can accurately reconstruct the car's size and the 2D location, while preserving the key perceptive features such as orientation and shape.

While the fog and rain particles do introduce additional environmental reflections in the radar heat-map, our GAN model is able to ignore environmental reflections and extract out only the key generative features for the car in the scene. 
\name's ability to create accurate images in fog and rain, despite never having been trained with such examples, demonstrates that we can leverage 
the favorable propagation characteristics of mmWave signals in inclement weather conditions to build a model that can generalize well between different environments and weather conditions. 


\vskip 4 pt
\noindent
{\bf Failure Examples:} Fig.\ref{fig:failure} shows some failure cases of \name. A current limitation of our system is that its performance deteriorates when the scene has multiple cars (Fig.\ref{fig:failure}(ii)). While \name\ correctly identifies the number of cars in the scene, it makes errors in their shapes and relative locations. To address this, a potential future direction is to adopt a {\it Region Proposal Network}, similar to
\cite{RF-Pose3D}, where \name\ can first isolate the reflections from each car
in the scene, and then reconstruct the body of each car individually.

In Fig.\ref{fig:failure}(iii) and (iv) we show failure cases with fog and rain respectively. It is worth noting that we are constrained to build our experimental setup at 
the 60 GHz unlicensed spectrum, which suffers from higher attenuation from water particles compared to other frequencies in the mmWave band. 
We believe that implementing \name\ at the 77 GHz band, which is specifically allocated for automotive radar applications, will result in even better 
performance. \name\ could also be specifically trained with images collected in bad weather to further improve the results.

\color{black}



\subsection{Quantitative Metrics}
\label{sec:quant}
\colorize 

We leverage a perception module to extract the following quantitative metrics that correspond to the contextual and perceptual information in the scene as described below: 
\vskip 4pt

\noindent
{\bf Ranging:} We compare the ranging performance of \name\ and radar to the ground truth. We assign the ranging distance to be the distance between the closest corner of the car and the radar imaging system in the 2D plane (Fig.~\ref{fig:metrics}(i)). We use the closest corner in our evaluation since the corners of the car scatter mmWave signals, and therefore, they can almost always be detected by the mmWave radar despite specular reflections from other parts of the car.

\vskip 4pt
\noindent
{\bf Size of Car: } We evaluate the performance on estimation of size by comparing the accuracy in length, width and height of the car.
For the 3D radar heat-map, we can measure the car dimensions as observed from the radar reflections. For the output from \name, we can measure the dimensions by projecting the 2D depth-map into a 3D point cloud.

\vskip 4pt
\noindent
{\bf Orientation of Car: } We measure the orientation as the angle between the longer edge of the car and the geographic north as viewed in the top view shown in Fig.~\ref{fig:metrics}(i). We can project the 2D depth-map or the 3D heat-map along the top view of the scene using standard rotational transformations.

\vskip 4pt
\noindent
{\bf Shape of Car: } We evaluate the shape of the car by comparing the boundary of the car as viewed along the front-view, that is, along the view of the 2D depth-map (Fig.~\ref{fig:metrics}(ii)). We consider the following two metrics:

\begin{itemize}[leftmargin=*,topsep=3pt]
	\setlength{\itemsep}{0pt} 
\setlength{\parskip}{0pt}

\item {\it Percentage of the Car Surface Missed ---} 
Indicative of the specularity effects observed in the image.


\item {\it Percentage of Fictitious Reflectors ---} Indicative of artifacts such as multipath and ambient relfections in the image.
\color{black}


\end{itemize}

\subsection{Quantitative Results}
\colorize 

In Fig.~\ref{fig:bar} we plot bar graphs depicting the performance of \name\
and the mmWave radar on the metrics described in the previous subsection. In
Fig.~\ref{fig:bar}(i) we plot the results for the real test dataset, whereas in
Fig.~\ref{fig:bar}(ii) we plot the results for the synthetic test dataset. As
can be observed, the results are similar for both and hereafter we will only discuss the results on the real dataset.

\vskip 4pt
\noindent
{\bf Ranging:} In Fig.~\ref{fig:bar}(i)(a), we plot the ranging error of the
car. Due to the high sensing bandwidth at mmWave frequencies, mmWave radars can
achieve high depth resolution. Therefore, mmWave radars can accurately detect
the closest corner of the car with a median error of 30 cm.
The skip connections in our design allow for direct
transfer of this ranging information from input to output, allowing \name\ to
accurately range the car with a 42 cm median error.

\vskip 4pt
\noindent
{\bf Size of Car:} From Fig.~\ref{fig:bar}(i)(b,c,d), we see that mmWave radar has a median error of
1.9 meters, 2.1 meters and 0.47 meters in length, width and height respectively. In comparison, \name\ achieves corresponding 
errors of 0.84 meters, 0.93 meters and 0.09 meters, offering a 2.26$\times$, 2.26$\times$ and 5.22$\times$ improvement
respectively. mmWave radars suffer from such large errors due to either specular reflections which lead to underestimation of the
car dimensions, or artifacts such as multipath and environmental reflections which results in overestimation. However, \name\ learns to fill in the holes in the radar heat-map to account for specularity and reject artifacts to only retain the true reflections from the car.

\vskip 4pt
\noindent
{\bf Orientation:} Fig.~\ref{fig:bar}(i)(e) shows the error in the estimated orientation of the car. 
As mentioned previously, we estimate the orientation from the top-view bounding box of the car. However, due to specularity and ghost reflections, the radar is unable to accurately estimate the bounding box and therefore leads to a large median error of 
$62^{\circ}$, whereas \name\ achieves a median error of $30^{\circ}$.

\color{black}

\vskip 4pt
\noindent
{\bf Shape of Car:} 
We evaluate the shape along two metrics:

\begin{itemize}[leftmargin=*,topsep=3pt]
	\setlength{\itemsep}{0pt} 
\setlength{\parskip}{0pt}

\item {\it Percentage of the Car Surface Missed ---} Fig.~\ref{fig:shape}(i)
		shows the CDF of the Percentage of the Car's surface missed by the
		radar and \name. This metric is similar to false negative
		rate with the 2D stereo camera depth-map as the ground
		truth. Due to specularity effects,
		most reflections from the car do not trace back to the receiver
		for the mmWave radar. As a result, mmWave radars miss a large
		portion of the car's surface, with the median value of
		Percentage Missed being $30\%$. In comparison,
		\name\ achieves a median value of $13\%$, thus showing that
		\name\ can effectively learn to fill in the holes in the image
		and can accurately capture the surface of the car.

\item {\it Percentage of Fictitious Reflectors ---} Fig.~\ref{fig:shape}(ii)
	shows the CDF of the Percentage of Fictitious Reflectors in the radar
		and \name image. This metric is similar to false positive
		rate. Radar heat-maps suffer from
		fictitious reflections due to artifacts such as multipath and
		environmental reflections, and as a result have a
		large number of false positives. In Fig.~\ref{fig:shape}(ii),
		we can see that the median value for percentage of fictitious
		reflections for the radar is $20\%$,
		whereas the $90\%$ percentile value is $38\%$. In comparison,
		\name\ achieves a median value of $3.7\%$ and a $90\%$
		percentile of $9.6\%$, resulting in an improvement of
		$5.5\times$ and $4\times$ respectively. This shows that \name\
		can effectively learn to reject the ghost reflections from the
		environment.

\end{itemize}

The above results show that \name\ faithfully captures the
shape of the car and reconstructs an accurate and high resolution image,
consequently outperforming state-of-the-art mmWave radar systems as demonstrated in this section.

	\section{Limitations and Discussion}
\label{sec:limit}

We have shown that \name\ is a promising approach for achieving high
resolution imaging with mmWave wireless systems, by leveraging the latest
advancements in AI and machine learning. However, \name\ has some limitations.

\begin{itemize}[leftmargin=*,topsep=3pt]
	\setlength{\itemsep}{0pt} 
\setlength{\parskip}{0pt}

	\item {\bf Real-time Implementation:} Our current implementation uses SAR to collect raw mmWave
	heat-maps. A more practical choice would be to use phased arrays that
		would enable realtime capture even as the cars are moving.
		However, this is infeasible due to the lack of cheap commercially available very large phased
		arrays today.
		As phased array technology becomes more accessible in the
		following years~\cite{Bell, UCSD}, we plan to transform \name\ to a realtime
		system that supports mobility. {{Note that Doppler shifts introduced by moving 
		cars in the scene will not affect \name's performance, since we can easily estimate and correct
		for Doppler shifts using our triangular FMCW waveform, as is standard in radar signal processing~\cite{Doppler}.}} 

\item {\bf Extending Beyond Cars:} 
	We train \name\ to specifically reconstruct cars.  
		However, a practical system should extend
		to image pedestrians, bicycles, traffic signs, etc. 
		One possibility is to adopt the approach from past
		work~\cite{RF-Pose3D} where a separate classification network
		first isolates reflections from each class of objects, and
		then employs a separate GAN model to reconstruct the shape of
		each object. There is also scope to further improve the
		performance of our system with multiple cars in the scene. 

	\item {\bf Even Higher Resolution:} Our current system offers an upscaling 
		factor of $4\times$ in the size of the image. However, 
		the actual increase in resolution is significantly higher
		as can be seen in the qualitative figures.  We could potentially
		have larger upscaling factors but this would
		result in an exponentially higher amount of resource
		requirement in terms of memory, compute power and size of
		the training dataset beyond what GPUs can currently handle.


\end{itemize}

\footnotesize
\bibliographystyle{ACM-Reference-Format}
\bibliography{related_bib}


\end{sloppypar}

\end{document}